\documentclass[lettersize,journal]{IEEEtran}
\pdfoutput=1
\usepackage{amsmath,amsfonts}
\usepackage{multirow,booktabs}
\usepackage{algorithmic}
\usepackage{array}
\usepackage[caption=false,font=normalsize,labelfont=sf,textfont=sf]{subfig}
\usepackage{textcomp}
\usepackage{colortbl}
\usepackage{stfloats}
\usepackage{url}
\usepackage{verbatim}
\usepackage{graphicx}
\usepackage{url}
\usepackage{amsmath,amssymb,latexsym}
\usepackage{graphicx}
\usepackage{hyperref}
\usepackage{cite}

\def\BibTeX{{\rm B\kern-.05em{\sc i\kern-.025em b}\kern-.08em
    T\kern-.1667em\lower.7ex\hbox{E}\kern-.125emX}}
\usepackage{balance}

\begin{document}
\title{DINO-CoDT: Multi-class Collaborative Detection and Tracking with Vision Foundation Models}
\author{Xunjie He* , Christina Dao Wen Lee*, Meiling Wang, Chengran Yuan, Zefan Huang, \\ Yufeng Yue \dag, Marcelo H. Ang Jr.
\thanks{*These authors contribute equally to this work.
}
\thanks{\dag Corresponding author: Yufeng Yue (yueyufeng@bit.edu.cn).
}
 \thanks{This work is partly supported by the National Natural Science Foundation of China under Grant 92370203, 62473050, 62233002.}
 \thanks{Xunjie He, Meiling Wang, Yufeng Yue are with School of Automation, Beijing Institute of Technology, Beijing, 100081, China.

Christina Dao Wen Lee, Chengran Yuan, Zefan Huang, Marcelo H. Ang Jr. are with Advanced Robotics Centre, Mechanical Engineering, National University of Singapore, Singapore.
 }}

\markboth{Journal of \LaTeX\ Class Files,~Vol.~18, No.~9, September~2020}%
{How to Use the IEEEtran \LaTeX \ Templates}

\maketitle

\begin{abstract}
Collaborative perception plays a crucial role in enhancing environmental understanding by expanding the perceptual range and improving robustness against sensor failures, which primarily involves collaborative 3D detection and tracking tasks. The former focuses on object recognition in individual frames, while the latter captures continuous instance tracklets over time. However, existing works in both areas predominantly focus on the vehicle superclass, lacking effective solutions for both multi-class collaborative detection and tracking. This limitation hinders their applicability in real-world scenarios, which involve diverse object classes with varying appearances and motion patterns.
To overcome these limitations, we propose a multi-class collaborative detection and tracking framework tailored for diverse road users. We first present a detector with a global spatial attention fusion (GSAF) module, enhancing multi-scale feature learning for objects of varying sizes. Next, we introduce a tracklet RE-IDentification (REID) module that leverages visual semantics with a vision foundation model to effectively reduce ID SWitch (IDSW) errors, in cases of erroneous mismatches involving small objects like pedestrians. We further design a velocity-based adaptive tracklet management (VATM) module that adjusts the tracking interval dynamically based on object motion.
Extensive experiments on the V2X-Real and OPV2V datasets show that our approach significantly outperforms existing state-of-the-art methods in both detection and tracking accuracy.

\end{abstract}


\begin{figure}
  \begin{center}
  \includegraphics[width=9cm]{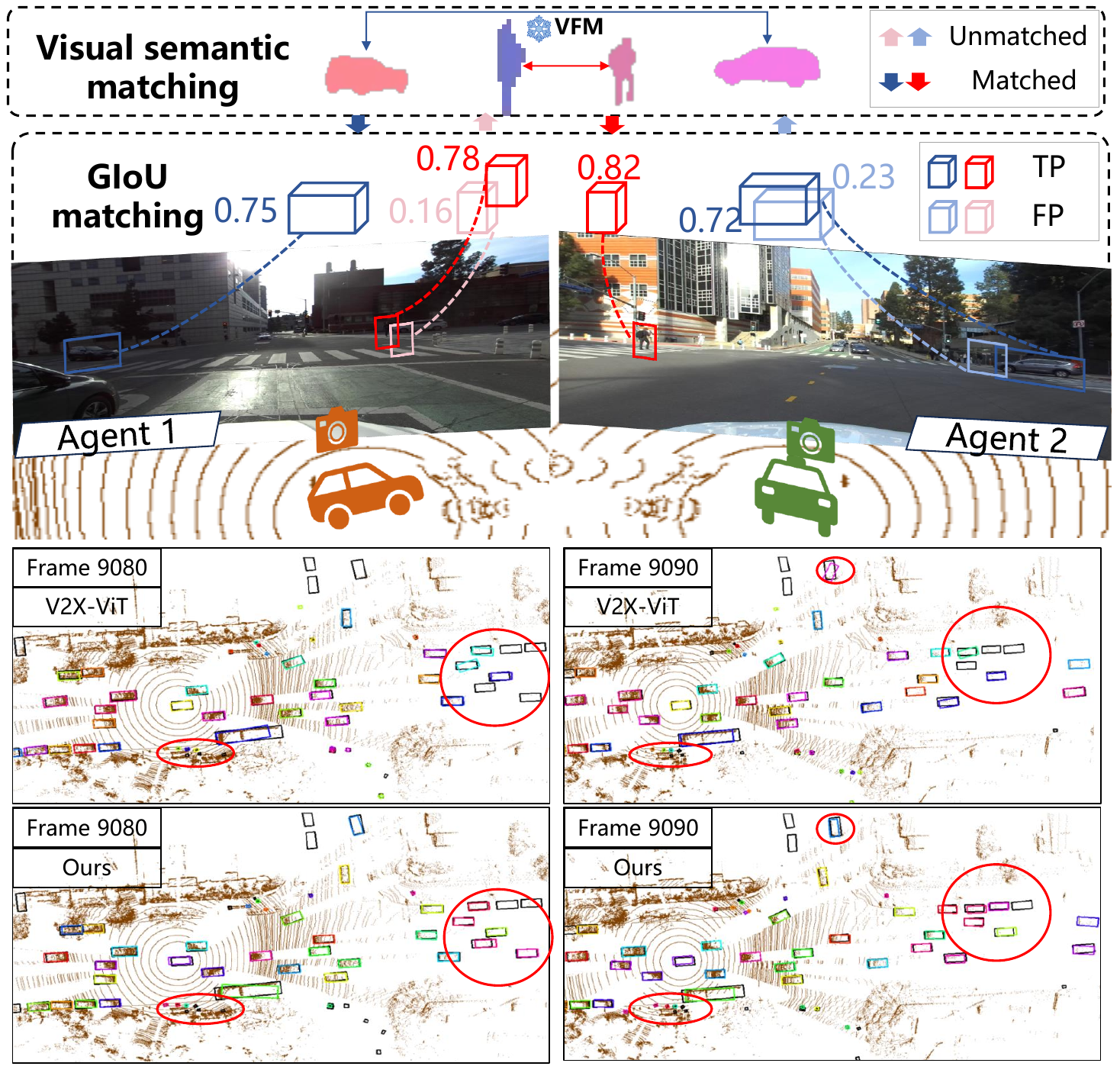}\\
  \caption{Comparative results of tracking in continuous frames. \textit{First row}: The visualization of DINO-CoDT highlights the specific design for tracking across multiple agents. \textit{Second \& Third row}: Tracking comparison results of V2XViT and ours.}\label{fig01}
  \end{center}
  \vspace{-0.8cm}
\end{figure}

\section{Introduction}
\IEEEPARstart{A}{utonomous} robotics technology has made significant advancements in the past decade. In this field, perception tasks serve as a critical foundation for subsequent mapping \cite{deng2024macim} and decision-making \cite{tang2023multi}, providing robust support for safe and efficient practical applications. Currently, an increasing number of researchers are focusing on collaborative perception \cite{10122468, chen2019f, xu2022v2x, he2024robust}. By integrating data from multiple agents, it effectively addresses the inherent limitations of single agent perception, such as limited field of view, restricted perception range, and susceptibility to occlusions. Multimodal collaborative datasets such as OPV2V \cite{xu2022opv2v} and V2X-Real \cite{xiang2024v2x} have also been gradually evolving and expanding, laying a solid foundation for real-world applications.

Collaborative perception in this work encompasses two fundamental perception tasks: collaborative 3D object detection and tracking. Most existing collaborative detection methods \cite{10382426, yin2023v2vformer++} focus on frame-level object recognition in static scenes. In contrast, collaborative tracking \cite{DMSTrack} aims to maintain consistent object identities across time and multiple agents, emphasizing multi-frame association and reasoning. However, multi-agent collaboration presents challenges such as bandwidth consumption and feature misalignment, making collaborative tracking less explored. Moreover, the presence of multiple classes in complex real-world environments introduces additional difficulties. Pedestrians, vehicles, and cyclists exhibit large variations in appearance, motion, and spatial behavior.  In a multi-agent setting, these differences become more challenging to handle due to inconsistent viewpoints, occlusions, and asynchronous observations across agents. However, most existing collaborative perception works \cite{xu2022v2x, DMSTrack} primarily target the vehicle superclass, limiting their ability to comprehensively understand diverse road users.
To address these challenges, we propose a multi-class collaborative perception framework (as shown in Fig. \ref{fig01}) that selectively incorporates information from multiple agents and introduces adaptive strategies to enable efficient and robust perception of diverse road users.
For collaborative detection, 
most existing benchmarks \cite{chen2019f, xu2022v2x} process data in the BEV perspective, where small objects such as pedestrians and certain types of vehicles are represented with fewer points, making them more difficult to detect. The disparities in object size and spatial distribution are further amplified in BEV representations, leading to increased perception errors.
To address the challenge of detecting vary-sized objects, we design a novel global spatial attention fusion (GSAF) module for inter-agent fusion, enhancing local context learning across all classes, particularly for pedestrian detection. This module dynamically explores the features of each individual object. For instance, it emphasizes fine-grained spatial details through multi-scale feature extraction, whereas in high-speed vehicle interactions, it captures broad contextual understanding via global attention mechanisms.

For collaborative tracking, most existing approaches \cite{li2024fast} rely heavily on geometric cues from 2D and 3D detections, which often leads to recognition errors and ID switches, especially in cases of occlusion or overlapping objects. Although LiDAR-based features dominate collaborative systems, they often fall short in capturing fine-grained semantics crucial for cross-agent association and multi-class differentiation. Recent advances in visual foundation models (VFMs) demonstrate powerful capabilities in extracting and aligning deep semantic features, offering a promising direction for addressing such challenges. To this end, we introduce a tracklet RE-IDentification (REID) module incorporating a vision foundation model DINOv2 \cite{oquab2024dinov2}. After extracting key features from cropped detection-tracking pairs, we compute similarity, and determine whether a tracked object and a detection represent the same entity, thereby recovering lost matches. In addition, motion prediction in current methods \cite{jung2024conftrack, AB3DMOT_eccvw} is typically based on traditional Kalman Filters with fixed parameters, such as birth and death thresholds, which are not well-suited to the diverse motion patterns of different object classes such as vehicles and pedestrians. To achieve consistent and accurate tracking, we implement a velocity-based adaptive tracklet management (VATM) mechanism for birth and death control, which optimizes the potential utility and leverages the maximum value of each object.

In summary, we present DINO-CoDT, a novel framework for collaborative detection and tracking in complex multi-class autonomous driving scenarios.Through extensive experiments on V2X-Real and OPV2V benchmark, DINO-CoDT demonstrates significant improvements in both detection and tracking accuracy. 



The main contributions of our work are as follows:
\begin{enumerate}
\item{We introduce DINO-CoDT, an innovative framework for multi-class collaborative detection and tracking. To the best of our knowledge, this is the first work that extends collaborative perception to support multiple object classes simultaneously.}
\item{We present a collaborative fusion strategy by introducing a global spatial attention fusion module to address the detection of objects at varying scales.}
\item{We propose a vision-based collaborative tracking approach that leverages the semantic association capabilities of a visual foundation model to reduce identity switches.}
\item{We design a velocity-based adaptive tracklet management method to accommodate the motion characteristics of different object classes.}
\end{enumerate}

The rest of this paper is organized as follows. Section II
outlines related works on collaborative 3D object detection and multi-object tracking. Section III provides an overview of the proposed network. Section IV introduces the
experimental details and evaluation of the proposed method. At last, Section V draws conclusions and discusses future work.

\section{Related Work}
\subsection{Collaborative 3D Object Detection}
Collaborative perception enhances multi-agent perception by fusing multi-view observations, overcoming single-agent limitations such as occlusion and limited perception range. The advancement of collaborative 3D object detection has primarily focused on two fundamental aspects: fusion strategies (Early/Intermediate/Late) and sensor modalities (LiDAR/Camera/Multimodal). Early fusion \cite{chen2019cooper, arnold2020cooperative} integrates raw sensor data, maintaining comprehensive environmental details at the expense of substantial bandwidth requirements. In response, late fusion techniques based on Non-Maximum Suppression (NMS) \cite{felzenszwalb2009object} are developed to exchange only final detection outputs, though this introduced error accumulation issues. Intermediate fusion \cite{wang2020v2vnet, liu2020when2com} emerges as the prevailing solution, which shared processed feature representations to optimally balance detection accuracy with system efficiency.
The selection of sensing modalities has equally influenced the field's progression. LiDAR-based systems \cite{chen2019f, xu2022opv2v, hu2022where2comm, liu2024select2col} deliver exceptional geometric precision but struggle with texture interpretation, while Camera-based approaches \cite{hu2023collaboration, xu2023cobevt} offer rich semantic information yet suffer from depth estimation limitations. Recent multimodal fusion architectures \cite{yin2023v2vformer++} aim to combine these complementary strengths, though challenges persist in achieving reliable spatiotemporal alignment.
Despite these advances, current BEV-based fusion frameworks exhibit notable limitations when handling challenging scenarios, particularly for vulnerable road users, such as pedestrians in dynamic environments. These shortcomings primarily originate from inadequate multiscale feature learning and insufficient adaptability to varying object sizes, which provides inspiration and guidance for our research.

\subsection{3D Multi-Object Tracking}

Tracking-by-Detection (TBD) \cite{AB3DMOT_eccvw, dino_mot} remains the dominant framework in Multi-Object Tracking (MOT). A key component of TBD is data association, which is typically based on metrics such as Intersection over Union (IoU), Euclidean distance, or appearance feature embeddings in methods like QDTrack \cite{QDTrack}. A representative method, AB3DMOT \cite{AB3DMOT_eccvw}, extends SORT \cite{SORT} into the 3D domain by leveraging LiDAR-based detections, employing a 3D Kalman filter and the Hungarian algorithm for association. While lightweight and effective, its application is largely limited to point cloud data and rarely applied to image-based tracking. To address this, DINO-MOT \cite{dino_mot} integrates pedestrian re-identification with visual tracking by utilizing foundation models, offering an image-centric solution. Building upon these foundations, Samba \cite{SambaMOT} constructively proposes that death models should account for occlusion dynamics, providing new insights into class-aware tracking strategies. Recently, collaborative tracking has begun to emerge as a promising direction. For instance, DMSTrack \cite{DMSTrack} improves association accuracy by modeling the observation noise covariance of each detection. However, its reliance on conventional collaborative datasets restricts its applicability to vehicle-only scenarios, limiting its generalization to more diverse and realistic environments.



\section{Methodology}
\subsection{Approach Overview}

 \begin{figure*}
\includegraphics[width=1\textwidth]{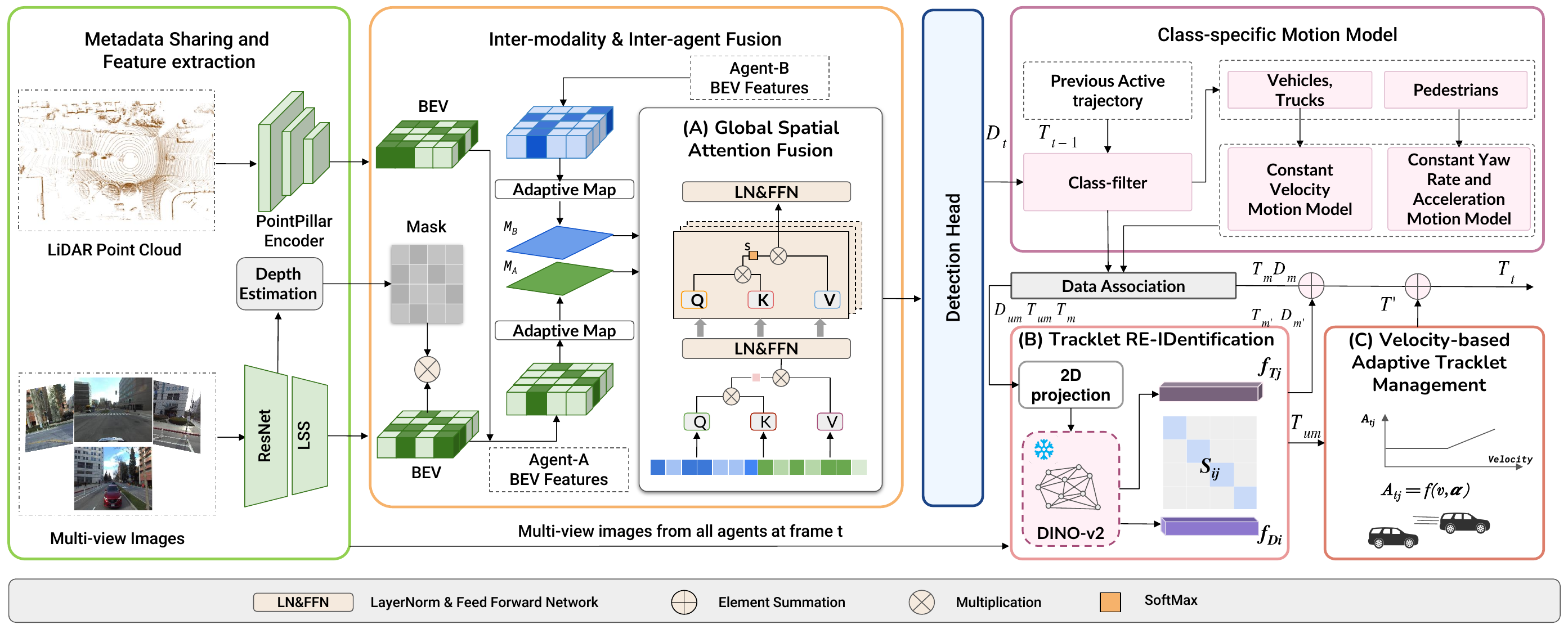}
\caption{Overall architecture of DINO-CoDT at frame \textit{t}. The framework contains three innovative modules: (A) Global Spatial Attention Fusion (GSAF) module in Section III-D, which works for effective multi-agent feature fusion. (B)Tracklet RE-IDentification (REID) module in Section III-G, which captures visual semantics with VFMs and completes association for unmatched pairs. (C)Velocity-based Adaptive Tracklet Management (VATM) module in Section III-H, which generates adaptive death threshold for objects with different velocity.}
\label{fig02}
\end{figure*}
We propose a comprehensive collaborative perception framework named DINO-CoDT, which is composed of two complete works: multi-class collaborative 3D detection and collaborative multi-object tracking. 
The collaborative 3D detection framework \cite{hu2022where2comm} comprises four key steps: metadata sharing, feature extraction, compression and fusion, and detection head. Considering \textit{N} agents in a collaborative system, the inputs in frame \textit{t} consist of point clouds $\textbf{P}^t_i, i \in [1, \dots, N]$ and four-view images of each agent $\textbf{I}^t_{ij}, j \in [1,2,3,4]$. After selecting a random agent as the designated ego agent, data from other agents are projected into the ego one's coordinate system. For different modalities, we extract features using suitable backbones. Following feature extraction, both image and point cloud features from multiple agents undergo intermediate compression and fusion, where we propose \textbf{a Global Spatial Attention Fusion (GSAF) module (details explained in Section III-D)} and generate a fused global feature. Finally, they are followed by supervised training using the current frame's label $\textbf{Y}^t_i$ to predict the final detection results. Furthermore, the detections undergo NMS and output $\textbf{D}_t$, which reduces false positives and increases the detection accuracy. 

In contrast to conventional single-agent MOT approaches, our framework leverages multi-agent multi-view images $\textbf{I}^t_{ij}$ to minimize tracking loss and mitigate ID switches. The architecture operates in a cyclic manner, utilizing trajectory predictions $\textbf{T}_{t-1}$ from the previous frame as partial inputs for the current frame. To accommodate diverse motion characteristics, we employ class-specific motion models built upon Kalman Filter for state estimation. The processed detections and predicted states are then integrated into a two-stage data association module, which generates robust ID matching pairs. Then the unmatched elements are fed into a \textbf{Tracklet RE-IDentification (REID) module (explained in Section III-G)}, which is enhanced by multi-view perspectives and facilitates tracking refinement in \textbf{Velocity-based Adaptive Tracklet Management (VATM) module (explained in Section III-H)}. The updated trajectories $\textbf{T}_{t}$ are subsequently propagated to the subsequent frame for continuous tracking. The overall architecture is shown in Fig. \ref{fig02}. Specific components and details are described in the sections below.

\subsection{Metadata Sharing and Feature Extraction}
For point cloud feature extraction, we adopt the PointPillar baseline due to its efficiency in extracting voxel features. Unlike point clouds, which inherently provide rich geometric and spatial information owing to their 3D nature, images require depth estimation to achieve precise transformation into a BEV representation. Depth estimation serves as a critical step, bridging the gap between 2D image data and 3D spatial understanding. To strike a balance between information richness and computational efficiency, we integrate a ResNet-based feature extractor with a GeneralizedLSS (Lift-Splat-Shoot) neck, inspired by the BEVFusion \cite{liu2023bevfusion} framework. Specifically, the ResNet backbone extracts high-level semantic features from the input images, while the GeneralizedLSS neck enhances depth estimation accuracy through a multi-scale feature fusion strategy. As a result, the geometric precision of point clouds and the semantic richness of image features are effectively captured, forming the foundation for a comprehensive representation tailored to downstream tasks.

\subsection{Inter-modality and Inter-agent Fusion}
During the feature compression phase, we employ multiple sequential $1 \times 1$ convolutional layers to compress BEV features along the channel dimension, simulating real-world communication constraints and reducing transmission bandwidth between agents. The intermediate fusion process is divided into inter-modality fusion and inter-agent fusion, addressing both multi-modal and multi-agent challenges. Before performing inter-modality fusion, we adopt a collaborative depth mask generation strategy that combines prediction and projection. Specifically, the image feature encoder also predicts pixel-wise depth from the camera view, which is then complemented by depth information projected from point clouds onto the 2D image, enhancing the overall depth estimation accuracy. To further refine the depth masks, point clouds from multiple agent viewpoints are projected and selectively fused. Based on the resulting voxels from both modalities and corresponding depth masks, a confidence-aware fusion is conducted using conditional convolutions. This enables the effective integration of point cloud and image information, resulting in more comprehensive and high-quality fused features. In the inter-agent fusion stage, the ego agent acts as the central node in a graph-based fusion network, while surrounding agents are treated as connected nodes. With the transformed features from connected agents, we design a global spatial attention fusion (Section III-D) module to produce a final fused output that captures both local details and global context. 

\subsection{Global Spatial Attention Fusion}

To effectively perceive objects across diverse scales, particularly small targets often missed in unified BEV representations, we propose a global spatial attention fusion framework that integrates both local and global contextual information in a scale-adaptive manner. Rather than treating each scale independently, our design promotes seamless information flow across multi-resolution representations, dynamically modulating feature interactions to suit varying object sizes. This mechanism is incorporated into the feature fusion module to effectively guide the model's attention towards semantically meaningful regions while preserving fine-grained details critical for small object detection.

With the intermediate feature maps $ \mathbf{F} \in \mathbb{R}^{N \times H \times W \times C} $ from N agents, we partition $ \mathbf{F} $ into $P$ scale-specific branches, generating a representation $ \mathbf{F}_s \in \mathbb{R}^{H_s \times W_s \times C} $ for each agent via non-overlapping windows of size $ w_s \times w_s $. Within each scale, local self-attention is performed to capture scale-specific spatial dependencies. And we compute attention for each scale $ s $ in branch $p$ as follows:
\begin{equation}
\mathbf{Z}_{p,s} = \psi_{h=1}^H \left( \text{SM} \left( \frac{\mathbf{Q}_s^{(h)} (\mathbf{K}_s^{(h)})^\top}{\sqrt{d_k}} + \mathbf{B}_s^{(h)} \right) \mathbf{V}_s^{(h)} \right) \mathbf{W}_o,
\end{equation}
where $\mathbf{Q}_s^{(h)}, \mathbf{K}_s^{(h)}, \mathbf{V}_s^{(h)} \in \mathbb{R}^{w_s^2 \times d_k} $ are the projected query, key, and value matrices for head $ h $, and $ \mathbf{B}_s^{(h)} $ denotes the relative positional bias. We use a SoftMax (SM) function to generate normalized weights. And a learnable linear projection $ \mathbf{W}_o \in \mathbb{R}^{(H \cdot d_k) \times C} $ is applied to the concatenated outputs from all attention heads, restoring the feature dimensionality back to $ C $. In addition, the matrix $ \mathbf{W}_o $ is trained jointly with the rest of the network. $\psi(\cdot)$ refers to a Concatenation function by which we concatenate two input features along the channel dimension.
After the multi-scale attention of each branch, we make a concatenation for them and output each agent's feature as $\mathbf{F}'= \psi_{p=1}^P\mathbf{Z}_{p,s}$.


For inter-agent feature fusion, we apply an attention mechanism by fusing each agent's feature and their adaptive weights:
\begin{equation}
\tilde{\mathbf{F}}^{(n)}_{i,j} = \sum_{m=1}^{N} \beta_{m\rightarrow n}(i,j) \cdot \mathbf{F}'^{(m)}_{i,j}
\end{equation}
where $ \beta_{m\rightarrow n}(i,j) \in [0,1] $ is the attention weight indicating the relevance of agent $ m $'s feature to agent $ n $ at position $ (i,j) $, computed as:
\begin{equation}
\beta_{m\rightarrow n}(i,j) = \frac{\exp\left(\text{MLP}([\psi(\mathbf{F}'^{(n)}_{i,j},\mathbf{F}'^{(m)}_{i,j})])\right)}{\sum_{k=1}^{N} \exp\left(\text{MLP}([\psi(\mathbf{F}'^{(n)}_{i,j},\mathbf{F}'^{(k)}_{i,j})])\right)},
\end{equation}
where $ \text{MLP}(\cdot) $ is a lightweight shared MLP used to calculate cross-agent similarity. This formulation allows each agent to adaptively incorporate complementary features from its peers at each spatial location, enhancing robustness to partial observations or occlusions.

Finally, the fused features are passed through a residual-enhanced feed-forward layer to form the output representation:
\begin{equation}
\mathbf{F}_{\text{out}} = LayerNorm(\tilde{\mathbf{F}} + \text{MLP}(\tilde{\mathbf{F}})), 
\end{equation}
where the normalization $LayerNorm(\cdot)$ and residual design stabilizes the feature distribution and enhances the expressivity of the final representations.

Our global spatial attention fusion module unifies multi-scale self-attention, cross-scale interaction, and inter-agent adaptive weighting, enabling the network to dynamically emphasize fine-scale features in regions with dense or small objects while leveraging coarse-scale features for larger structures. This design enhances the model’s ability to preserve small object boundaries and reduce spatial ambiguity, particularly in cluttered or occluded environments, ultimately improving both recall and precision in small object detection under collaborative perception settings.


\subsection{Detection Head}
The detection decoder processes the fused features to produce class and regression predictions. Specifically, the class output provides the predicted class and confidence score of an anchor box. The regression output defines a rotated bounding box $(x, y, z, w, l, h, \theta)$, where $(x, y, z)$, $(w, l, h)$, and $\theta$ denote the position, size and yaw angle, respectively. To enhance detection accuracy, we implement a confidence score filter and apply NMS to remove redundant boxes with significant overlaps, which is denoted as preprocessing before tracking.

\subsection{Class-specific Motion Model}
In order to capture the highly nonlinear motion features of objects of all classes, we make a class-specific motion module according to the different motion characteristics of different classes. For the classes like vehicles and trucks, we use a constant velocity motion model. For classes like pedestrians and bicycles, we design a constant turn rate and acceleration motion model. This considers the nonlinear property of the prediction to complete the state estimation $\hat{x}$ of the track object for different classes.

\begin{figure}
  \begin{center}
  \includegraphics[width=3.3in]{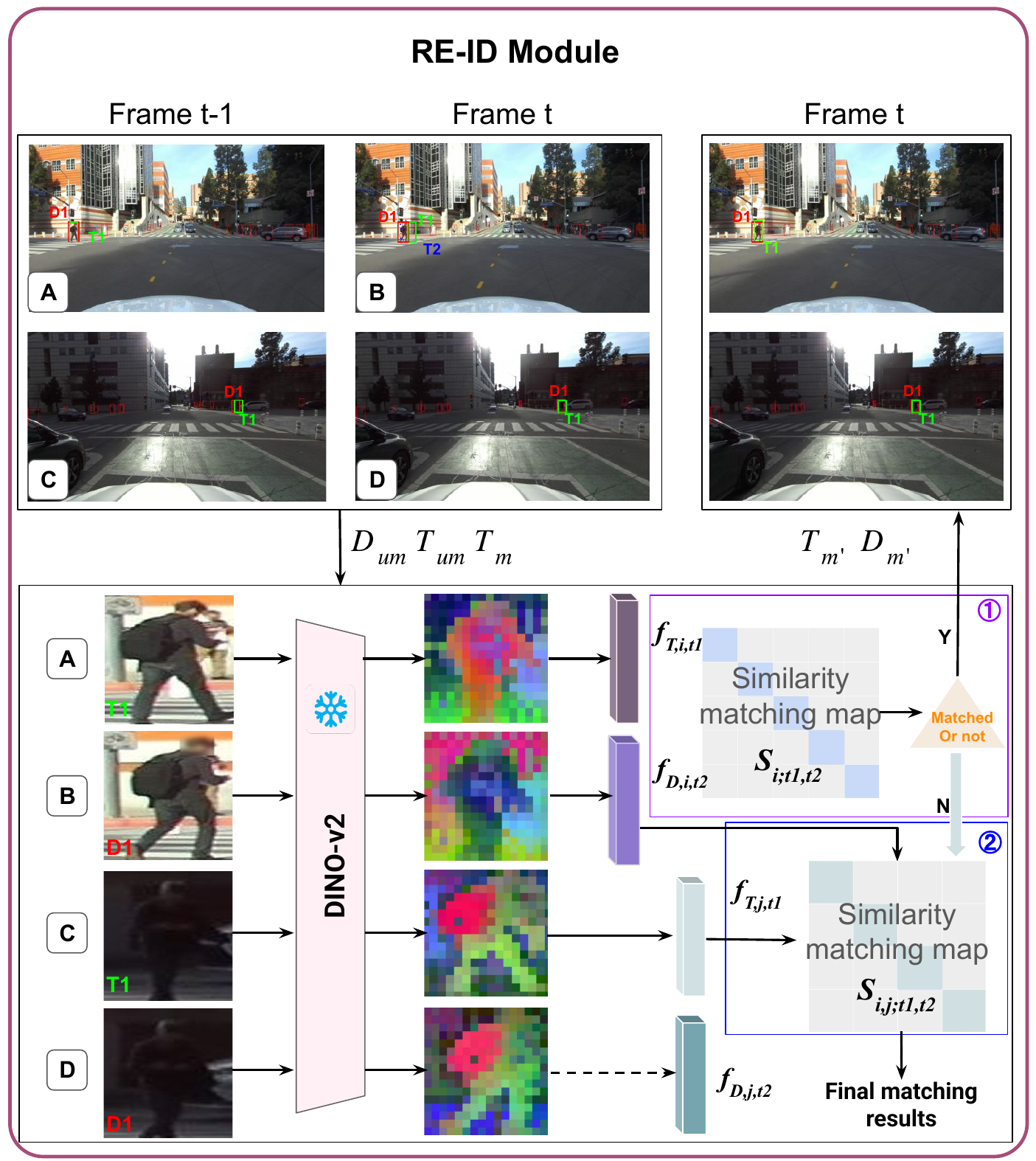}\\
  \caption{Pipeline of Tracklet RE-IDentification (REID) module. Our method utilizes a vision foundation model DINOv2 to compare semantic information of targets from multiple viewpoints across agents, enabling the correction of erroneous associations and enhancing matching reliability.}\label{fig:03-reid}
  \end{center}
\end{figure}

\subsection{Data Association and Tracklet RE-IDentification }
Given the predicted state of the tracked object $\textbf{T}_{t-1}$ and the detection $\textbf{D}_t$, we attain the affinity matrix by computing the 3D Intersection over Union (IoU) between every pair of existing tracks and detection, following the conventional approach used in AB3DMOT \cite{AB3DMOT_eccvw}. The 3D IoU quantifies the spatial overlap between predicted trajectories and detected objects. 
After constructing the affinity matrix, we formulate the data association task as a bipartite graph matching problem 
and employ the matching algorithm (either Hungarian or Greedy Algorithm \cite{li2023assignment}, based on the class of object), which optimally assigns detections to trajectories by maximizing the overall affinity score while ensuring that each detection is matched to at most one trajectory and vice versa. 

Following the data association, we implement a tracklet RE-IDentification module that generates new matched pairs by incorporating visual semantics from unmatched detections, ultimately producing the final tracklets. Our approach begins by projecting each detection onto raw multi-view images. Unlike single-agent imagery, multi-agent images provide comprehensive object descriptions from multiple perspectives, effectively compensating for detection losses in individual frames. Using camera intrinsics and the Ego-to-Camera transformation matrixs, we compute the position of the object in the image space shown in (\ref{eqn:transform}),(\ref{eqn:projection}). The bounding boxes are determined by the values ($u_{min}, v_{min}, u_{max}, v_{max}$ ) of these coordinates, which define the object's 2D projection for region cropping.

\begin{equation}
\begin{bmatrix}
x_{\text{c}} \\
y_{\text{c}} \\
z_{\text{c}} \\
1
\end{bmatrix}
=
^{a} T_{\text{ego}}  \cdot
\begin{bmatrix}
x_{\text{ego}} \\
y_{\text{ego}} \\
z_{\text{ego}} \\
1
\end{bmatrix} ,
\quad \text{where } a \in \{N_{Agents}\}
\label{eqn:transform}
\end{equation}

\begin{equation}
\begin{bmatrix}
u \\
v \\
1 
\end{bmatrix} 
=
\mathbf{K}^a \cdot
\begin{bmatrix}
x_{\text{c}} \\
y_{\text{c}} \\
z_{\text{c}} 
\end{bmatrix} ,
\quad \text{where } a \in \{N_{Agents}\}
\label{eqn:projection}
\end{equation} 


For visual semantic extraction, we adopt a state-of-the-art vision foundation model \cite{oquab2024dinov2} to capture rich and discriminative object representations as shown in Fig. \ref{fig:03-reid}. Specifically, inspired by AnyLoc \cite{keetha2023anyloc}, we utilize the 31st encoder layer of DINOv2 to extract high-dimensional semantic embeddings for each detected region. Given an input image $I$ and a bounding box corresponding to the current detection $D_r$ or an existing track $T_r$, we first apply a cropping function $\phi_{crop}(\cdot)$ to isolate the region of interest. The cropped image patch is then forwarded through the DINOv2 backbone, and the output is taken as the semantic embedding:
\begin{equation}
    f_{rd} =\mathcal{F}^{(31)}_{DINOv2}(\phi_{crop}(I,D_r)), f_{rd} \in \mathbf{R}^d,
\end{equation}
\begin{equation}
    f_{rt} =\mathcal{F}^{(31)}_{DINOv2}(\phi_{crop}(I,T_r)), f_{rt} \in \mathbf{R}^d,
\end{equation}
where $\mathcal{F}^{(31)}_{DINOv2}(\cdot)$ denotes the feature extraction function at the 31st transformer block of DINOv2, and $f_{rd}$ and $f_{rt}$ are the resulting feature vector in a high-dimensional space $\mathbf{R}^d$. This embedding enables efficient cross-object comparison without the need for domain-specific fine-tuning, thereby ensuring both robustness and computational efficiency.

Following the DINO-MOT paradigm, we maintain a dynamic Look-Up Table (LUT) that stores feature vectors $f_{rt}$ indexed by their corresponding track IDs. 
For each incoming detection, we compute the cosine similarity between its visual embedding and historical features within a predefined temporal window shown in (\ref{eqn:cosine_similarity}). 
\begin{equation}
\text{Cosine Similarity}(f_{\text{rd}}, f_{\text{rt}, \beta}) = \frac{f_{\text{rd}} \cdot f_{\text{rt}}}{\|f_{\text{rd}}\| \|f_{\text{rt}}\|},
\label{eqn:cosine_similarity}
\end{equation}
where a strict visual similarity threshold of $\beta$ is established to ensure matching precision, where exceeding this value triggers a track association event.

In scenarios where no confident match is found within the temporal window, we further extend the matching scope to a collaborative spatial domain by leveraging cross-agent information. Specifically, each unmatched detection is additionally compared with the feature memories of other agents at the same timestamp. Using the same similarity metric, we identify potential inter-agent correspondences and perform a secondary association if the cross-agent similarity surpasses a predefined inter-agent threshold. This inter-agent matching strategy complements the temporal matching by enabling robust identity propagation even in cases of partial occlusion, viewpoint change, or sensor failure within a single agent. 

During this process, the system performs track ID correction through a memory update mechanism: the historical track ID supersedes the current detection's ID, while simultaneously refreshing the feature bank with the latest visual representation. 
This approach effectively maintains temporal consistency while adapting to appearance variations.

\subsection{Velocity-based Adaptive Tracklet Management}
Tracked objects of different classes exhibit distinct motion patterns and behaviors, so it is essential to design a flexible mechanism to manage the birth (entry) and death (exit) of objects adaptively. For instance, vehicles typically move at higher speeds and follow predictable trajectories, while pedestrians may exhibit slower and more erratic movements. To address this, we propose a velocity-based adaptive mechanism that dynamically adjusts death thresholds at time $t$ for tracklet $j$. The adaptive max age $A_{tj}$ is computed based on the observed motion characteristics of each object class as (\ref{age}):
\begin{equation}
\label{age}
A_{tj} = A_c + \alpha \cdot \sqrt{v_x^2 + v_y^2 },
\end{equation}
where $v_x$ and $v_y$ are velocity in the x and y direction respectively, taken from the state estimate of the tracklet $j$. $\alpha$ is the weight applied to the velocity. And $A_c$ is the original max age for each class $c$.
 
\begin{table*}[!htbp]
\caption{Specific class list in three superclasses in V2X-Real dataset.}
\centering
\label{superclass}
\renewcommand\arraystretch{1.3}
\begin{tabular}{c|c|ccc|cc}
\toprule
Superclass & Vehicle & \multicolumn{3}{c|}{Pedestrian} & \multicolumn{2}{c}{Truck} \\
\midrule
\multirow{3}*{Classes} & Long Vehicle & Child & RoadWorker & Pedestrian & Truck  & Van\\
 & Car & Scooter & ScooterRider & Motorcycle & TrashCan & ConcreteTruck \\
& PoliceCar & MotorcycleRider & BicycleRider & & Bus & \\
\bottomrule
\end{tabular}
\end{table*}

\section{Experiments Results}
In this section, the proposed DINO-CoDT algorithm is evaluated with extensive experiments on both simulated and real-world collaborative perception datasets.

\subsection{Datasets}

\subsubsection{V2X-Real Dataset}
V2X-Real \cite{xiang2024v2x} is a large-scale real-world dataset tailored for V2X collaborative perception research. Specifically, the whole dataset contains 33K LiDAR frames, 171K camera images and over 1.2M 3D bounding box annotations. Each frame is recorded at a frequency of 10Hz, and $23379$, $2770$, and $6850$ frames are contained in the training, validation and test sets, respectively. The perception range is set as $x \in [-100m, 100m]$, $y \in [- 40m, 40m]$, $z \in [- 15m, 15m]$.

This dataset establishes the first multi-class collaborative benchmark, encompassing a diverse range of vulnerable road users such as pedestrians, scooters, motorcycles, and bicycles, which are systematically classified into three superclasses (Vehicle, Pedestrian and Truck) based on their bounding box dimensions. Each superclass is divided into specific classes as explained in Table \ref{superclass}. Its distinctive infrastructure integrates two smart road-side units and two autonomous vehicles within each frame, organized into four specialized subdatasets according to ego-agent types and their collaborative interaction patterns. Our current research focuses on the V2X-Real-V2V configuration as a representative case, and the proposed framework is inherently extensible to other subdataset configurations.

\begin{table*}[!htbp]
\caption{Collaborative detection performance comparison on the V2X-Real \cite{xiang2024v2x} dataset. The baseline methods' results are from the V2V collaborative perception benchmark and the results are reported in AP@0.3 and AP@0.5. The best results are in BOLD.}
\centering
\label{detection}
\renewcommand\arraystretch{1.3}
\begin{tabular}{c|cc|cc|cc|cc}
\toprule
Classes  & \multicolumn{2}{c}{Vehicle} & \multicolumn{2}{|c|}{Pedestrian} & \multicolumn{2}{c|}{Truck} & \multicolumn{2}{c}{Avg}\\
\cmidrule(lr){2-9}
Network/Metric & AP@0.3 & AP@0.5 & AP@0.3 & AP@0.5 & AP@0.3 & AP@0.5 & AP@0.3 & AP@0.5\\
\midrule
No fusion & 41.7 & 39.4 & 26.9 & 14.4 & 21.6 & 13.7 & 30.1 & 22.5 \\
Late fusion & 47.4 & 44.4 & 29.2 & 14.9 & 18.7 & 9.1 & 31.8 & 22.8 \\
Early fusion & 54.0 & 49.8 & 31.9 & 17.1 & 28.6 & 18.6 & 38.1 & 28.5\\
F-Cooper \cite{chen2019f} & 42.7 & 40.3 & 27.7 & 14.0 & 25.6 & 18.6 & 32.0 & 24.3 \\
AttFuse \cite{xu2022opv2v} & 58.6 & 55.3 & 30.1 & 15.4 & 28.9 & 21.7 & 39.2 & 30.8\\
V2X-ViT \cite{xu2022v2x} &  59.0 & 56.3 & 37.4 & 20.7 & 42.9 & 35.0 & 46.5 & 37.3\\
\midrule
Ours &  \textbf{68.7} & \textbf{66.0} & \textbf{41.5} & \textbf{22.2} & \textbf{49.6} & \textbf{48.5} & \textbf{53.3} & \textbf{45.6} \\
\bottomrule
\end{tabular}
\end{table*}

\subsubsection{OPV2V Dataset \cite{xu2022opv2v}}
OPV2V is a large-scale dataset simulated for agent-to-agent collaborative perception. It features over 70 diverse scenarios collected from 8 unique towns in CARLA and a highly detailed digital reconstruction of Culver City, Los Angeles, totaling 11,464 frames and 232,913 accurately annotated 3D vehicle bounding boxes.
Each frame typically involves around 3 connected autonomous vehicles (CAVs), each equipped with a $360^{\circ}$ camera system (comprising 4 cameras), a 64-channel LiDAR, and GPS/IMU sensors. The data is recorded at 10 Hz.
The perception range is defined as $x \in [-140\,\text{m}, 140\,\text{m}]$, $y \in [-40\,\text{m}, 40\,\text{m}]$, and $z \in [-3\,\text{m}, 1\,\text{m}]$. The dataset is split into 6,765 frames for training, 1,980 for validation, and 2,720 for testing, with the test set including 2,170 frames from the Default Town and 550 frames from Culver City.

\subsection{Implementation Details}
\subsubsection{Detection}
Prior to training, each voxel is configured with a width and length of both 0.4 meters, and a height of 4 meters. We employ the Adam optimizer with a weight decay of 0.0001. The learning rate is initially set to 0.0001 and follows a multi-step learning rate (MultiStepLR) scheduling strategy, where it decays by a factor of 0.1 at epochs 10 and 20. Our model focuses on detecting three superclasses and we evaluate our work on the test set. The training process is conducted over 50 epochs using a single NVIDIA RTX 3090 GPU.

\subsubsection{Tracking} 
In DINO-CoDT, we perform Tracking by Detection (TBD). Detections are first filtered with a confidence threshold, only detections with a confidence level exceeding 0.1 are considered; this is applied to all classes. 

We implement multi-object tracking in a collaborative fashion, using camera images from all connected agents to perform re-identification. We tune the maximum age of all classes with adaptive threshold $\alpha$ and IoU threshold as explained in Section IV-E-2. For the motion model, Vehicle and Pedestrian classes follow a Constant Yaw Rate and Acceleration (CYRA) motion model, whereas Truck follows a Constant Velocity model.

\subsection{Evaluation Metrics}
 For detection, the annotated 16 classes are grouped into three super classes i.e. vehicle, pedestrian and truck as per their bounding box size distribution. We employ the same detection evaluation metrics from V2X-Real \cite{xiang2024v2x} for each class: Average Precision (AP) at an Intersection-over-Union (IoU) threshold of 0.3/0.5.
A final mean Average Precision (mAP) is also evaluated based on each class’s AP:
\begin{equation}
     mAP=\frac {1}{C}\sum _{i=1}^{C}AP_i,
\end{equation}
where C is the class number.

In object tracking tasks, we use the same evaluation metrics as AB3DMOT to measure the performance of tracking algorithms. They are scaled Average Multi-Object Tracking Accuracy (sAMOTA), Average Multi-Object Tracking Accuracy (AMOTA), Average Multi-Object Tracking Precision (AMOTP), Multi-Object Tracking Accuracy (MOTA), Multi-Object Tracking Precision (MOTP), and the number of ID SWitches (IDSW). Specifically, MOTA is a comprehensive metric that evaluates tracking accuracy  in each sequence. It is calculated as:
\begin{equation}
    MOTA=1 - \frac{FP+FN+IDSW}{GT},
\end{equation}
where $FP$ is the number of false positives, $FN$ is the number of false negatives, and $GT$ is the number of ground truth objects.
MOTP measures the precision of object localization, typically computed as the average distance between predicted and ground truth object positions. It is defined as:
\begin{equation}
    MOTP= \frac{\sum_{t=1}^T\sum_{i=1}^{M_t}o_{t,i}}{\sum_{t=1}^T M_t},
\end{equation}
where $T$ is the total number of frames, $M_t$ is the number of matched objects at frame $t$, and $o_{t,i}$ is the intersection over union between the predicted and ground truth positions of the matched object $i$ at frame $t$.

\begin{table*}[!htbp]
\caption{Collaborative tracking performance comparison on the V2X-Real \cite{xiang2024v2x} dataset. The baseline results are evaluated by feeding the detection results to the AB3DMOT method \cite{AB3DMOT_eccvw}. Ours* represents the tracking evaluation by feeding our detection results into AB3DMOT.The best results are in BOLD, and the second best results are \underline{underlined}.}
\centering
\label{tracking}
\renewcommand\arraystretch{1.3}
\resizebox{16cm}{!}{
\begin{tabular}{p{1.7cm}<{\centering}p{0.7cm}<{\centering}p{0.5cm}<{\centering}p{0.5cm}<{\centering}p{0.7cm}<{\centering}|p{0.7cm}<{\centering}p{0.7cm}<{\centering}p{0.5cm}<{\centering}p{0.7cm}<{\centering}|p{0.7cm}<{\centering}p{0.6cm}<{\centering}p{0.5cm}<{\centering}p{0.7cm}<{\centering}}
\toprule
Classes & \multicolumn{4}{c}{Vehicle} & \multicolumn{4}{|c|}{Pedestrian} & \multicolumn{4}{c}{Truck}\\
\cmidrule(lr){2-13}
Network/Metric & $\scriptstyle \text{sAMOTA} \scriptstyle \uparrow$
 & $\scriptstyle \text{AMOTA} \scriptstyle \uparrow$
 & $\scriptstyle \text{AMOTP} \scriptstyle \uparrow$
 & $\scriptstyle \text{MOTA} \scriptstyle \uparrow$
 & $\scriptstyle \text{sAMOTA} \scriptstyle \uparrow$ 
 & $\scriptstyle \text{AMOTA} \scriptstyle \uparrow$
 & $\scriptstyle \text{AMOTP} \scriptstyle \uparrow$
 & $\scriptstyle \text{MOTA} \scriptstyle \uparrow$
 & $\scriptstyle \text{sAMOTA} \scriptstyle \uparrow$
 & $\scriptstyle \text{AMOTA} \scriptstyle \uparrow$
 & $\scriptstyle \text{AMOTP} \scriptstyle \uparrow$
 & $\scriptstyle \text{MOTA} \scriptstyle \uparrow$ \\
\midrule
F-Cooper \cite{chen2019f} & 37.60 & 8.44 & 26.18 & 28.87 & 5.01 & -3.91 & 13.18 & 2.80 & 7.82 & 0.50 & 13.37 & 7.22 \\
AttFuse \cite{xu2022opv2v} & 36.67 & 8.03 & 27.80 & 27.19 & 16.59 & 0.43 & \underline{19.28} & 9.32 & 6.61& -1.74& 15.20 & 4.93 \\
V2X-ViT \cite{xu2022v2x} & 48.45 & 13.72 & 36.50 & 37.70 & 18.37 & 3.33 & 18.83 & 12.99 & 8.91 & -0.73 & 23.52 & 5.50\\
\midrule
Ours* & \underline{49.19} & \underline{14.34} & \underline{39.78} & \underline{39.06}  & \underline{22.98} & \underline{4.15} & 18.63 & \underline{14.67} & \underline{20.43} & \underline{3.28} & \underline{25.51} & \textbf{12.96} \\
Ours & \textbf{50.68} & \textbf{15.33} & \textbf{41.29} & \textbf{39.56} & \textbf{23.75} & \textbf{4.46} & \textbf{19.74} & \textbf{14.91}  &   \textbf{22.89} & \textbf{3.85} & \textbf{27.13}	& \underline{12.66}  \\
\bottomrule
\end{tabular}}
\end{table*}

 \begin{table*}[!htbp]
\caption{Comparison results of detection and tracking on the OPV2V dataset \cite{xu2022opv2v}. The best results are in bold. }
\centering
\label{opv2v}
\renewcommand\arraystretch{1.3}
\resizebox{14cm}{!}{
\begin{tabular}{c|cc|ccccc}
\toprule
 & \multicolumn{2}{c|}{Detection} & \multicolumn{5}{c}{Tracking}\\
\cmidrule(lr){2-8}
Network/Metric & AP@0.5 & AP@0.7 
 & $ \text{sAMOTA} \uparrow$
 & $ \text{AMOTA}  \uparrow$
 & $ \text{AMOTP} \uparrow$
 & $ \text{MOTA}  \uparrow$
 & $ \text{MOTP}  \uparrow$

\\
\midrule
F-Cooper \cite{chen2019f} & 84.6 & 76.7 & 55.30 & 17.44 & 49.99 & 43.19 & 69.69 \\
AttFuse \cite{xu2022opv2v}& 82.4 & 67.2 & 57.42 & 18.35 & 47.02 & 43.97 & 66.85 \\
V2X-ViT \cite{xu2022v2x}& 85.9 & 77.5 & 51.90 & 17.72 &  49.03 & 44.61 & 66.80 \\
Ours* & \textbf{89.8} & \textbf{81.6} & 56.93 & 19.38 & 52.50 & 47.18 & \textbf{70.32} \\
Ours &\textbf{89.8} & \textbf{81.6} & \textbf{57.98} & \textbf{20.34} & \textbf{53.97} & \textbf{48.20} & 69.97 \\
\bottomrule
\end{tabular}}
\end{table*}

\subsection{Quantitative Comparison}
\subsubsection{Detection Comparison}
To the best of our knowledge, this is the first work to accomplish multi-class collaborative detection by effectively integrating both point cloud and image data. Considering that multi-class algorithms are still relatively scarce in the field of collaborative perception, we adopt the representative baseline methods from the V2X-Real dataset for comparative evaluation, such as No fusion, Late fusion, Early fusion, F-Cooper \cite{chen2019f}, AttFuse \cite{xu2022opv2v} and V2X-ViT \cite{xu2022v2x}. Among these methods, No fusion is considered as the baseline, which only uses individual observation. Late fusion shares the detected 3D bounding boxes with nearby agents while Early fusion fuses the raw data from all agents. As shown in Table \ref{detection}, our method achieves notable performance gains across all object classes compared to the original dataset benchmark. Even with the inclusion of the pedestrian superclass, the detection performance still exhibits significant improvement, demonstrating the effectiveness of our approach in multi-class scenarios. The substantial improvements can be attributed to the introduction of global spatial attention fusion. By capturing fine-grained local information, this module substantially improves the model’s capacity to detect objects of different scales from the BEV perspective, leading to average accuracy gains of 6.8\% at AP@0.3 and 8.3\% at AP@0.5.

\begin{table*}[!htbp]
\caption{Ablation analysis on variants of the whole collaborative perception system. Due to space constraints, this table focuses on evaluating the detection and tracking performance of the pedestrian class. A is a GSAF module. B is a REID module. C is a VATM module. The best results are in BOLD. }
\centering
\label{ablation}
\renewcommand\arraystretch{1.3}
\resizebox{16cm}{!}{
\begin{tabular}{ccc|cc|ccccccccc}
\toprule
\multicolumn{3}{c}{Network/Metric} & \multicolumn{2}{|c|}{Detection} & \multicolumn{9}{c}{Tracking}\\
\cmidrule(lr){1-14}
A & B & C & AP@0.3 & AP@0.5 
 & $ \text{sAMOTA} \uparrow$
 & $ \text{AMOTA}  \uparrow$
 & $ \text{AMOTP} \uparrow$
 & $ \text{MOTA}  \uparrow$
 & $ \text{MOTP}  \uparrow$
 & $ \text{IDSW}  \downarrow$
 & $ \text{TP}  \uparrow$
 & $ \text{FP}  \downarrow$
 & $ \text{FN}  \downarrow$\\
\midrule
 \multicolumn{3}{c|}{Baseline}& 37.4 & 20.7 &  18.37 & 3.33 & 18.83 & 12.99  & 46.18 & 4016 & 16734 & \textbf{5479} & 39003\\
 \midrule
 \checkmark &  & & \multirow{4}{*}{\textbf{41.5}} & \multirow{4}{*}{\textbf{22.2}} & 22.98 & 4.15 & 18.63 & 14.67  & 47.15 & \textbf{3316} & 18109 & 6614 & 37628\\
 & \checkmark & & & & 18.55 & 3.40 & \textbf{19.98} & 13.12 & 46.18 & 3927 & 16724 & 5482 & 39013 \\
 &  & \checkmark & & & 18.53 & 3.42 & 19.95 & 12.91 & 46.14 & 4046 & 16723 & 5480 & 39014 \\
 \checkmark  & \checkmark  & \checkmark &  & & \textbf{23.75} & \textbf{4.46} & 19.74 & \textbf{14.91} & \textbf{47.20} & 3746 & \textbf{19499} & 7445 & \textbf{36238} \\
\bottomrule
\end{tabular}}
\end{table*}

\subsubsection{Tracking Comparison}
We present a comprehensive comparison of our tracking performance against existing methods from established benchmarks. To ensure a fair evaluation, the detection outputs from all approaches are processed using the standard tracking algorithm AB3DMOT \cite{AB3DMOT_eccvw}. Concretely, we denote the combination of our detector with AB3DMOT as `Ours*' in the table, which serves as a strong baseline.
Remarkably, our detector, when paired with AB3DMOT, already surpasses almost all competing methods across all object classes, also underscoring the strong discriminative capability and reliability of our detection framework. Building upon this, our proposed tracking method, DINO-CoDT (denoted as `Ours'), further improves performance over the AB3DMOT-based baseline as shown in Table \ref{tracking}. This consistent gain highlights the superiority of our tracking algorithm and its robustness in handling diverse object classes and motion patterns in collaborative perception scenarios.

\subsubsection{Generalization evaluation}
As a supplementary experiment, we further evaluate our approach on the widely-used simulated OPV2V dataset. Although OPV2V only includes vehicle superclass such as Cars and SUVs, it still serves as a valuable benchmark for assessing the generalization capability and robustness of our method compared to existing approaches.
As shown in Table \ref{opv2v}, our method achieves notable improvements in detection performance, with gains of at least 3.9\% at AP@0.5 and 4.1\% at AP@0.7 over baseline methods. In addition, our tracking framework demonstrates consistent improvements across nearly all evaluation metrics, further validating the effectiveness of our detection and tracking pipeline even in scenarios with limited object diversity. These results reinforce the adaptability of our approach across both real-world and simulated datasets.

\begin{figure*}[!ht]
  \begin{center}
  \includegraphics[width=7in]{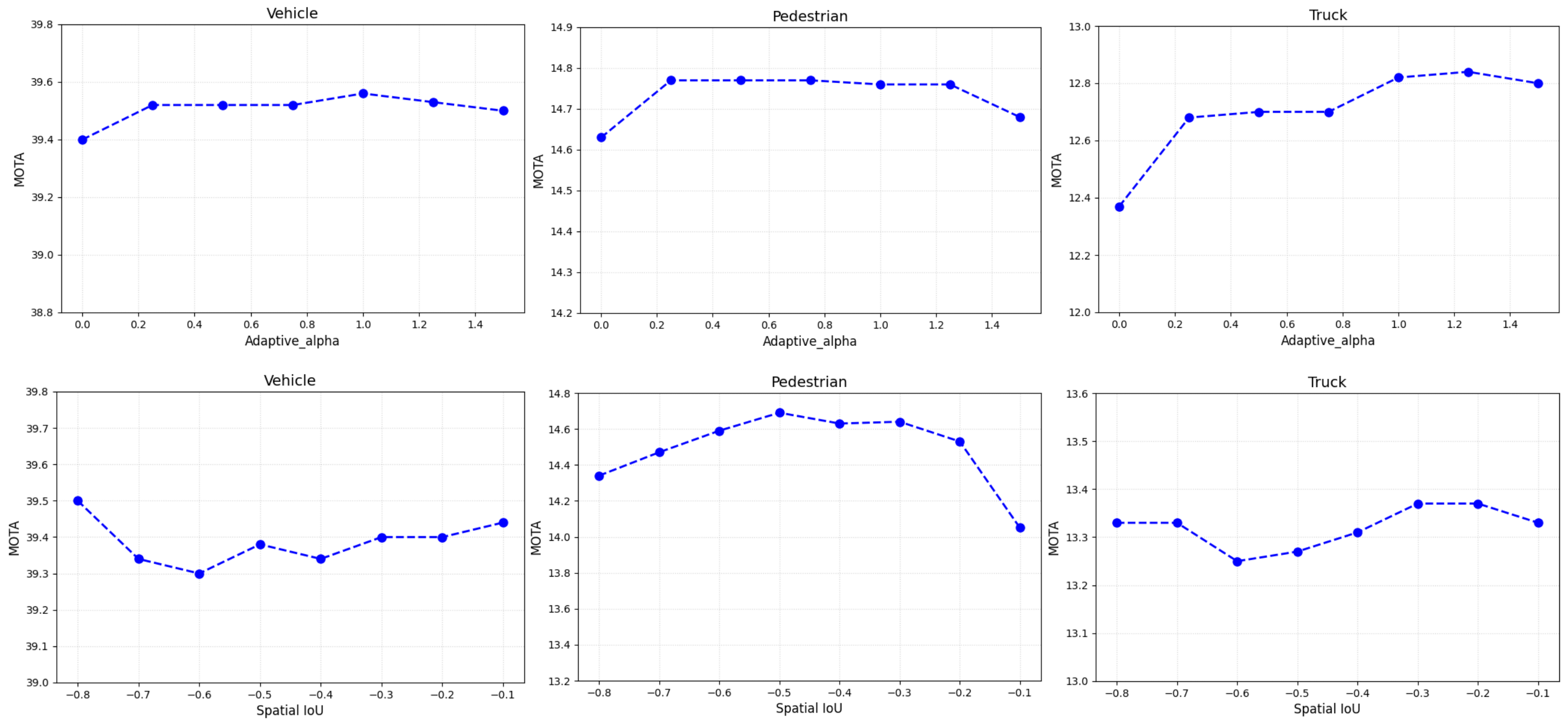}\\
  \caption{The qualitative comparison results of ours under different hyperparameters on the V2X-Real dataset \cite{xiang2024v2x}. The adaptive threshold $\alpha$ and IoU threshold are mentioned in the VATM module.}\label{hyperparameter}
  \end{center}
\end{figure*}

\subsection{Ablation Study}
\subsubsection{Inserted Modules}
Table \ref{ablation} presents the results of our ablation study, highlighting the individual contributions of each proposed module within our framework. We adopt V2X-ViT \cite{xu2022v2x} as the baseline and integrate our modules respectively to evaluate their impact. 

\textbf{Module A} refers to the enhancements made to the detection pipeline, which is specifically a proposed Global Spatial Attention Fusion (GSAF) module based on an extended multi-modal fusion framework. These improvements result in a substantial boost in tracking performance across nearly all evaluation metrics. Notably, we observe a 4.61\% increase in sAMOTA and a 1.68\% gain in MOTA, clearly indicating the effectiveness of the enhanced detector in multi-agent collaborative settings.

\begin{table}[!htbp]
\caption{Comparison results of common image encoder ResNet and DINOv2 for the superclass Pedestrian on the V2X-Real dataset. The best results are in bold. }
\centering
\label{dinov2}
\renewcommand\arraystretch{1.5}
\begin{tabular}{p{1.2cm}<{\centering}|p{1.4cm}<{\centering}|p{0.7cm}<{\centering}p{0.7cm}<{\centering}p{0.7cm}<{\centering}p{0.7cm}<{\centering}p{0.5cm}<{\centering}}
\toprule
Network/ Metric & Feature Extractor  & $ \scriptstyle \text{sAMOTA} \scriptstyle \uparrow$
 & $\scriptstyle \text{AMOTA} \scriptstyle \uparrow$
  & $\scriptstyle \text{AMOTP} \scriptstyle \uparrow$
 & $\scriptstyle \text{MOTA} \scriptstyle \uparrow$  & $\scriptstyle \text{IDSW} \scriptstyle \downarrow$\\
\midrule
no REID & No & 22.98 & 4.15 & 18.63 & 14.67  & 3316 \\
\midrule
\multirow{3}{*}{With REID} & ResNet-18 & 23.16 & \textbf{4.35} & 19.80 & 14.52 & 3823\\
 ~ & ResNet-34 & 22.98 & 4.30 & \textbf{19.81} & 14.49 & 3382\\
 ~ & ResNet-50 & 23.32 & 4.32 & 19.79 & 14.48 & 3867\\
 ~ & DINOv2 & \textbf{23.43} & 4.34 & 19.76 & \textbf{14.53} & \textbf{3301} \\
\bottomrule
\end{tabular}
\end{table}

\textbf{Module B} investigates the incorporation of DINOv2 for tracklet RE-IDentification (REID). With this addition, we observe a slight yet meaningful improvement of 0.18\% in sAMOTA, alongside a reduction in ID SWitches from 4016 to 3927. This demonstrates the module’s capability to reduce identity mismatches and maintain track consistency over time, especially in crowded or occluded environments.

To further validate the strong semantic extraction capability of the visual foundation model, we conduct a more detailed ablation study focusing on the visual encoder within the ReID module. As shown in table \ref{dinov2}, we choose classic ResNet-based image feature extractors, which are ResNet-18, ResNet-34 and ResNet-50, to make a comparison. As a result, the application of DINOv2 leads to a significant performance improvement, which further demonstrates the critical role of semantic understanding of vision foundation models.

\textbf{Module C} evaluates the impact of our Velocity-based Adaptive Tracklet Management (VATM) module. While the overall improvement appears marginal with a 0.16\% increase in sAMOTA compared to the baseline, it is important to note that the limited effect may stem from the relatively low speed and movement variance of dense pedestrians, which increases the necessity for velocity-aware management strategies in such cases.

The complete framework, integrating all the proposed modules, achieves an accuracy improvement of 5.38\% in sAMOTA and 1.92\% in MOTA. This experiment focuses on the most challenging pedestrian superclass, fully demonstrating the broadening of the perception capabilities of existing algorithms for complex road users.

\begin{figure*}[!ht]
  \begin{center}
  \includegraphics[width=7in]{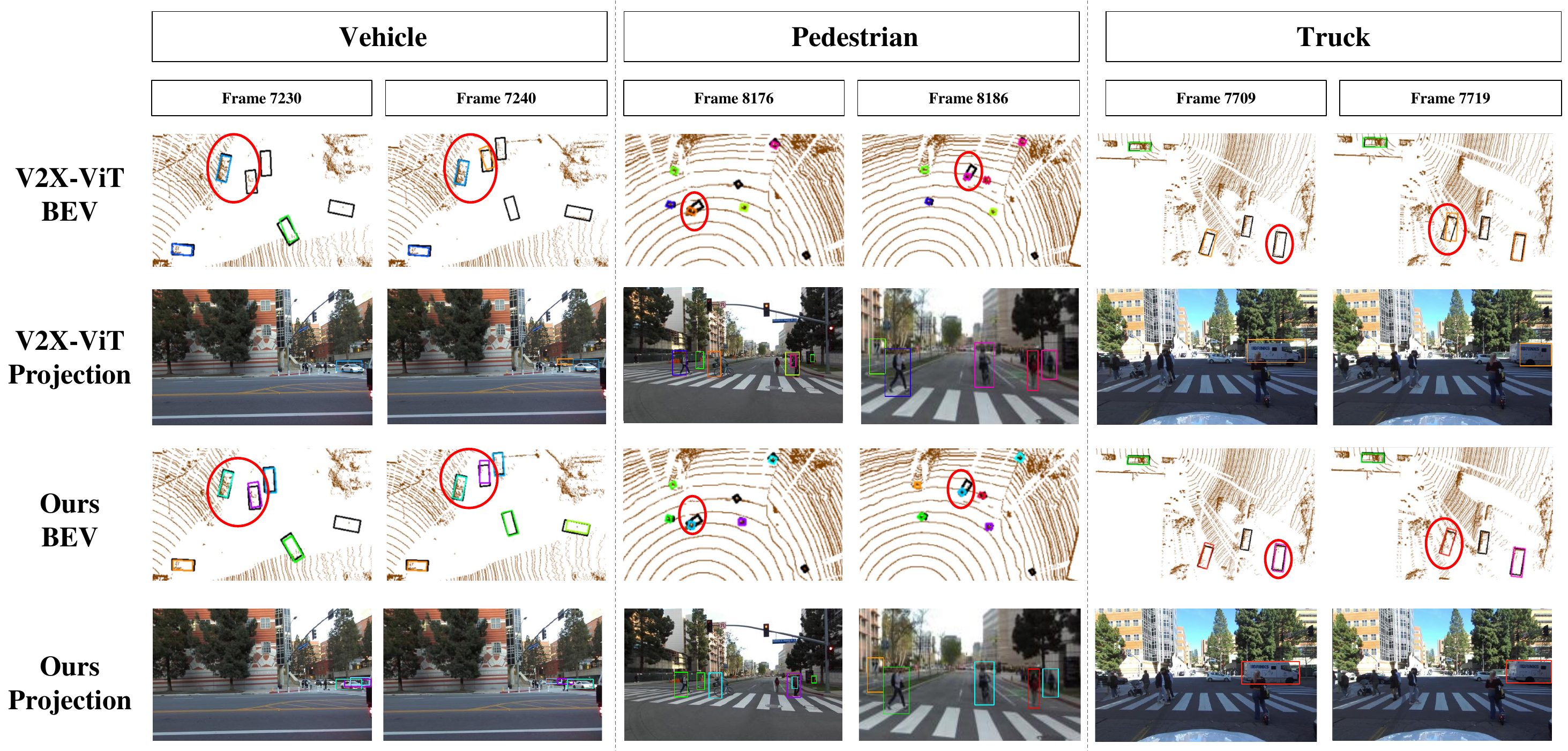}\\
  \caption{The qualitative comparison results of different methods on the V2X-Real dataset \cite{xiang2024v2x}. Tracking targets with different IDs are assigned distinct colors. The crucial regions are highlighted in red circles.}\label{qualitative}
  \end{center}
\end{figure*}


\subsubsection{Effects of Hyperparameters}
To investigate the influence of crucial hyperparameters on the optimal performance of our algorithm, we conduct comparative experiments focusing on how specific hyperparameters affect both class-level performance and overall accuracy. In particular, we analyze the impact of the velocity-adaptive threshold $\alpha$ proposed in VATM, as well as the overlap IoU threshold used for matching. As illustrated in Fig \ref{hyperparameter}, increasing adaptive threshold ($\alpha >0$) generally leads to improved MOTA, with performance peaking around 1.0–1.2 for most classes. This effect is most pronounced for Truck, while Pedestrian shows a relatively stable response, indicating that moderate adaptation enhances tracking performance across the board. In contrast, variations in the IoU threshold lead to more pronounced fluctuations in performance. Specifically, MOTA improves within a certain threshold range (approximately -0.5 to -0.2) and declines outside it. If set too low, it results in a large number of false positives; if set too high, it leads to excessive false negatives. Both scenarios increase the likelihood of mismatches and degrade tracking performance. The Pedestrian category is particularly sensitive to this parameter, suggesting the existence of an optimal spatial overlap threshold. In summary, both these two hyperparameters contribute positively to tracking accuracy when properly tuned, and their influence is broadly consistent across different object classes, underscoring their importance in multi-object tracking optimization.

\subsection{Qualitative Comparison}
In this paper, we conduct a qualitative comparison between our proposed method and the baseline V2X-ViT \cite{xu2022v2x}. We evaluate tracking performance across three object classes, each assessed over continuous frames featuring visual transformations, and present results from both the BEV and projected 2D image perspectives. As illustrated in Fig. \ref{qualitative}, our method consistently delivers superior tracking results across most superclasses, offering more accurate and coherent object trajectories.
For example, in the pedestrian superclass, the baseline method frequently suffers from identity switches, as highlighted in the red circle. In contrast, our approach effectively addresses these challenges through the integration of tracklet re-identification, maintaining more stable identities over time. Compared to the baseline, our method significantly reduces tracking failures and spatial misalignments, further demonstrating its robustness and effectiveness in multi-class and multi-agent scenarios.

\section{Conclusion}
In this work, to the best of our knowledge, we are the first to propose a comprehensive framework for multi-class collaborative detection and tracking. A GSAF module is introduced to enhance detection accuracy across objects of varying sizes through local multi-scale feature learning. To reduce identity mismatches during tracking, we design a REID module that explores semantic associations using a vision foundation model (DINOv2). Additionally, a VATM module is developed to adjust hyperparameters based on object motion characteristics, effectively mitigating tracking loss caused by detection errors. In all, our proposed work, DINO-CoDT shows improved performance on the V2X-Real dataset with +6.8\% on the AP@0.3 metric and +5.38\% on the sAMOTA metric compared to the existing method. This showcases its effectiveness in improving overall detection and tracking accuracy. 

Although our framework expands the range of recognizable object classes, it still relies on ground-truth labels. Future work can consider improving the detection network by incorporating advanced vision-language models for open-vocabulary perception. Moreover, multi-agent collaboration improves spatial awareness, but it also introduces high communication costs, which can be further optimized through compression, scheduling strategies, or decentralized architectures.

\bibliographystyle{IEEEtran}
\bibliography{Bibliography}

\end{document}